\documentclass[10pt,twocolumn,letterpaper]{article}

\usepackage{authblk}
\usepackage{iccv}

\usepackage{graphicx}
\usepackage{amsmath}
\usepackage{amssymb}
\usepackage{booktabs}

\usepackage{algorithmic}

\usepackage[ruled]{algorithm2e}
\usepackage{booktabs}
\usepackage{makecell}
\usepackage{multirow}
\usepackage{graphicx}
\usepackage{amsmath}
\usepackage{amssymb}

\newcommand{\myparagraph}[1]{\vspace{0.1em}\noindent\textbf{#1}}
\usepackage[pagebackref=true,breaklinks=true,letterpaper=true,colorlinks,bookmarks=false]{hyperref}

\usepackage[capitalize]{cleveref}
\crefname{section}{Sec.}{Secs.}
\Crefname{section}{Section}{Sections}
\Crefname{table}{Table}{Tables}
\crefname{table}{Tab.}{Tabs.}

\iccvfinalcopy % *** Uncomment this line for the final submission

\def\iccvPaperID{8783} % *** Enter the ICCV Paper ID here

\ificcvfinal\fi

\begin{document}

%%%%%%%%% TITLE
\title{ClickSeg: 3D Instance Segmentation with Click-Level Weak Annotations}

\author[1]{Leyao Liu}
\author[2]{Tao Kong}
\author[2]{Minzhao Zhu}
\author[1]{Jiashuo Fan}
\author[1]{Lu Fang}
\affil[1]{Tsinghua University}
\affil[2]{ByteDance}

\maketitle
% Remove page # from the first page of camera-ready.
\ificcvfinal\thispagestyle{empty}\fi

\begin{abstract}
3D instance segmentation methods often require fully-annotated dense labels for training, which are costly to obtain. 
In this paper, we present ClickSeg, a novel click-level weakly supervised 3D instance segmentation method that requires one point per instance annotation merely. 
% This paper aims to tackle the problem of point cloud instance segmentation with click-level weak supervision, which only requires the annotation of one point per instance, thus significantly reducing the annotation efforts.  
%To address the problem of extremely limited supervisions, we propose a self-training method.
%, we consider generating pseudo instance labels for the training set.
%%%%%%
% We first introduce a baseline self-training method that performs on 3D instance segmentation, then we give specific improvements under the click-level annotation setting. In detail, we utilize the property of such annotation setting that in each scene, every unlabeled point must belong to one of the annotated points. Instead of directly utilizing the model inference way, i.e., mean-shift clustering, to generate pseudo labels, we propose to use k-means clustering with fixed initial seeds: the annotated points. Furthermore, we modify the similarity metrics to better adapt to weak supervision setting.
%%%%%
%%%%% new
Such a problem is very challenging due to the extremely limited labels, which has rarely been solved before. We first develop a baseline weakly-supervised training method, which generates pseudo labels for unlabeled data by the model itself. To utilize the property of click-level annotation setting, we further propose a new training framework. Instead of directly using the model inference way, i.e., mean-shift clustering, to generate the pseudo labels, we propose to use k-means with fixed initial seeds: the annotated points. New similarity metrics are further designed for clustering.
% Specifically, instead of directly using the model inference way, i.e., mean-shift clustering to generate pseudo labels, we treat it as a classification problem that assign each unlabeled points to one of the annotated points. we propose to use k-means with fixed initial seeds: the annotated points. We also define new similarity metrics used in clustering.
%\fang{rewrite:::::::Furthermore, utilizing the property of click-level annotation settings, we propose improved techniques which largely increase the accuracy.}
%%%%%
%\kt{To fully utilize the extremely limited supervisions, we propose the online pseudo instance label generating framework which gradually forms more accurate supervision signals.We assign each unlabeled points to one of the annotated instance points, and define similarity metrics to further refine them with fixed initial seeds:the annotated points.During training, the similarity metrics are simultaneously learned with the whole end-to-end framework.}
% This is treated by us as a classification problem that assigns each unlabeled points to one of the annotated points. We define similarity metrics and formulate the classification as k-means clustering with fixed initial seeds:
% the annotated points.
%use k-means clustering with initial seeds being the labeled points to classify the unlabeled points.
% The similarities are in turn further learned after the pseudo label generation. 
%The whole process can be seen as a self-training. 
% At inference phase, we directly use these similarity metrics to generate the final instance segmentation results. 
Experiments on ScanNetV2 and S3DIS datasets show that the proposed ClickSeg surpasses the previous best weakly supervised instance segmentation result by a large margin (e.g., +9.4\% mAP on ScanNetV2). Using 0.02\% supervision signals merely, ClickSeg achieves $\sim$90\% of the accuracy of the fully-supervised counterpart. Meanwhile, it also achieves state-of-the-art semantic segmentation results among weakly supervised methods that use the same annotation settings.
\end{abstract}

\section{Introduction}
\begin{figure}[t]
\centering
\includegraphics[width=0.38\textwidth, angle=0]{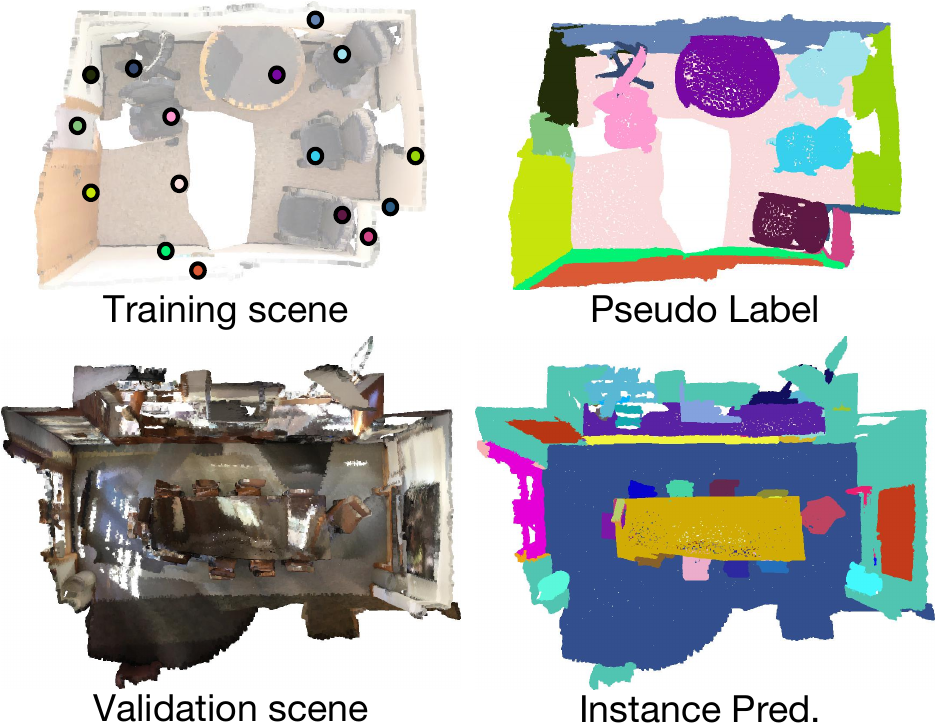}
\vspace{5pt}
\caption{ClickSeg only requires the annotators to annotate one point per instance and can generate fine quality instance segmentation results.}
\vspace{-10pt}
\label{fig:match_error}
\end{figure}
While 3D point cloud semantic and instance segmentation have achieved remarkable results in recent years, they often require dense supervision, which brings a great human burden. On average, a scene in ScanNet \cite{Dai2017ScanNetR3} contains 150k points and takes 22.3 minutes to annotate, while there are a total number of 1513 scenes. The situation is more severe in some datasets \cite{armeni2017joint, Hua2016SceneNNAS} that contain millions of points in a single scene. This largely hampers the application of fully-supervised algorithms.

Some recent methods \cite{xu2020weakly, liu2021one, wei2020multi} use weak labels to train the 3D semantic segmentation model and generate pseudo labels, but they mainly focus on semantic segmentation. Instance segmentation, however, is more complicated and hard to achieve using limited labels since it also requires the separation of all instances. Recently, Tao~\cite{tao2022seggroup} propose a method for this task. They first recover the complete training set by gradually merging supervoxels to their neighboring labeled ones, then perform fully-supervised methods. However, the training set is fixed once recovered, which contains many errors, thus their accuracy largely falls behind fully-supervised methods. Besides, they require the annotators to annotate the most central supervoxel of each instance, which is hard for humans to realize.
%The accuracy also largely falls behind fully-supervised methods.

In this paper, we present ClickSeg, a weakly supervised 3D instance segmentation method that only requires the annotators to annotate one arbitrary point per instance. We first develop a weakly-supervised training method that performs on 3D instance segmentation. It initially trains the model on weak labels by metric learning, then generates pseudo labels for unlabeled data by the inference way of the model. To fully utilize the property of click-level annotation setting, we further propose a new training framework. Since each instance is annotated for one point, every unlabeled point must belong to one of the annotated points. Thus, instead of directly using the model inference way, i.e., mean-shift clustering, to generate pseudo labels, we propose to use k-means clustering with fixed initial seeds: the annotated points. Furthermore, we define new similarity metrics used in clustering to better adapt to weak supervision setting, which helps pseudo label generation.
The pseudo labels are generated online during training, which are increasingly accurate.
This novel framework significantly improves the final accuracy.
% The similarities are first learned in supervoxels. 
%\kt{The pseudo labels could be gradually learned during the network training to further boost the model performance.}
% After generating the pseudo labels, they are then further learned. 
%So the pseudo label generation and training benefit from each other. 
%The whole process can be seen as a self-training. 

Experimental results show that ClickSeg surpasses the previous best weakly supervised 3D instance segmentation method by a large margin, and achieves $\sim$90\% of the accuracy of the fully supervised counterpart. Meanwhile, it also achieves state-of-the-art semantic segmentation results among other weakly supervised methods that use the same annotation settings.
To summarize, our contributions include:
\begin{itemize}
	\item  An effective and efficient 3D instance segmentation training framework which only requires click-level annotations. It generates pseudo labels online during training and the training time is almost the same as fully-supervised methods.\vspace{-5pt}
	\item A novel pseudo instance label generation method specifically designed for click-level annotation setting, which utilize k-means with fixed initial seeds: the annotated points.\vspace{-5pt}

	\item Extensive experiments demonstrate our superior performance over other weakly-supervised methods in both instance and semantic segmentation.
\end{itemize}

\section{Related Work}
\subsection{Point Cloud Segmentation}
3D point cloud segmentation is a fundamental computer vision task that has been widely studied. Recent methods could be mainly divided into two categories: point-based and voxel-based, according to the convolution type. The point-based methods \cite{Qi2017PointNetDH, Wu2019PointConvDC, Thomas2019KPConvFA, Qi2017PointNetDL} use point convolution to extract deep per point features. Voxel-based methods like \cite{cciccek20163d, Hou20193DSIS3S} first voxelize the point cloud, then perform 3D CNN on the grids. SparseConv \cite{Graham20183DSS} is later proposed to reduce the computation and memory consumption of these methods. % We adopt the same network backbone as it. 

Compared with semantic segmentation methods, instance segmentation methods face a more challenging situation, as they are also required to separate different instances.
% The semantic segmentation directly train using cross-entropy loss, while the 3D instance segmentation methods being more complicated. 
Current fully-supervised 3D instance segmentation methods can be categorized into proposal-based and grouping-based paradigms. 
The proposal-based methods use a top-down strategy to first generate the region proposals, then perform segmentation inside the region \cite{Hou20193DSIS3S,yang2019learning}. The grouping-based methods use a bottom-up scheme that group points to form the instances. We follow this kind of scheme and mainly introduce these methods below. 
%SGPN~\cite{wang2018sgpn} constructs a similarity matrix of all points and cluster them into instances according to their similarity. 
MTML~\cite{Lahoud20193DIS} outputs an embedding and a directional vector for each point. They use the embedding to cluster points and the directional vector to score the proposals. Occuseg \cite{Han2020OccuSegO3} proposes an occupancy signal to guide the clustering process. PointGroup \cite{Jiang2020PointGroupDP} uses semantics to group original and offset-shifted points, and then uses ScoreNet to score the proposals. HAIS \cite{chen2021hierarchical} follows this method and further absorbs small fragments into instances. SSTNet \cite{liang2021instance} constructs a superpoint tree to form instances via merge and split. These methods achieve remarkable results but require fully-annotated labels, which are difficult to obtain.

\begin{figure*}[t]
\centering
\includegraphics[width=0.9\textwidth, angle=0]{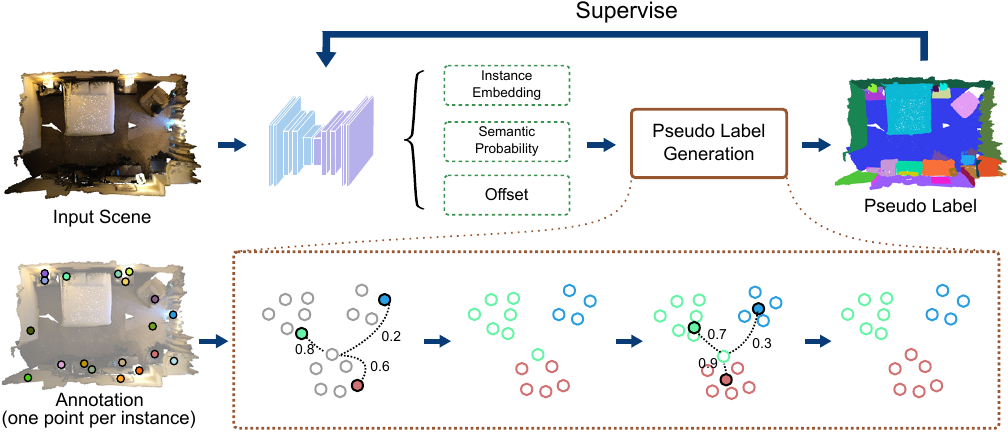}
\caption{The pipeline of ClickSeg. The network outputs several terms for each point of the input scene. We generate pseudo labels by using these terms as similarity metrics to perform k-means clustering with initial seeds being the annotated points. The network is in turn trained using these pseudo labels. Please refer to Sec.~\ref{sec:method} for more details.}
\label{fig:overview}

\end{figure*}

\subsection{Weakly Supervised Segmentation}
Weakly supervised image semantic segmentation have been extensively studied, including image-level \cite{ahn2018learning}, click-level \cite{bearman2016s}, scribble-level \cite{lin2016scribblesup}, and box-level \cite{dai2015boxsup} supervisions. The image-level supervision only indicates the contained classes of an image. Most of these methods \cite{wang2020self, huang2018weakly} use class activation map (CAM) to generate pixel-level labels. What's the point \cite{bearman2016s} is the first to introduce click-level supervision that annotates one point per object. It uses objectness prior, image-level and click-level supervision to form the loss function. ClickSeg uses the same annotation strategy. The scribble-level supervision is proposed in \cite{lin2016scribblesup} to draw a curve for each object. The box-level supervision uses bounding boxes to train a segmentation model. 

For weakly supervised 3D segmentation, the recently emerged methods mainly focus on semantic segmentation. MPRM \cite{wei2020multi} generates point-level labels from sub-cloud-level supervision using CAM. Xu \etal \cite{xu2020weakly} propose a method that only requires 10\% of labels to reach comparable results to their fully-supervised counterpart. SQN \cite{hu2021sqn} shows that very few labels are enough for semantic segmentation and propose a method that requires only 1\% labels for training. One-Thing-One-Click \cite{liu2021one} uses click-level labels and proposed a self-training semantic segmentation method that utilizes CRF for label propagation. Weakly supervised 3D instance segmentation, however, is an under-explored area. Recently, SegGroup \cite{tao2022seggroup} is proposed to tackle this problem. They require the annotator to annotate the most central supervoxel of each instance. They first recover the training set by merging neighboring supervoxels according to the features from graph convolution, and then perform fully-supervised training. 
% However, the training set contains many errors, resulting in low accuracy, and their annotation setting is difficult to achieve.

% self-training scheme has been widely used in weakly and semi-supervised learning methods \cite{rosenberg2005semi, liu2021one, yalniz2019billion, tarvainen2017mean, xie2020self}. It first trains the model on labeled data, then generates the pseudo labels using the model predictions on unlabeled data, and further trains on high confident unlabeled data. The process iterates until no new pseudo labels reach the predefined confidence thresh. 

\section{Method}
\label{sec:method}

We first introduce the fully supervised baseline method, which could be seen as the upper bound. We adopt a common fully supervised method \cite{Han2020OccuSegO3, zhang2020spatial} that uses a network backbone to output for each point $i$ an embedding vector $e_i$, a semantic probability $s_i$, and an offset vector $o_i$ pointing to the center of its belonging instance.
%then use clustering to group points to instances by their embeddings and offsets.
The backbone is the widely used 3D UNet implemented by Submanifold Sparse Convoliton \cite{Graham20183DSS}, which is shown to be effective with efficiency.

The semantic probabilities and the offset vectors are trained using cross-entropy loss $L_{CE}$ and L2 regression loss $L_{regress}$, respectively. The embeddings are trained by metric learning \cite{kaya2019deep}. Specifically, discriminative loss \cite{de2017semantic} is used, which reduces the distances of embeddings from the same instance by using the loss function of:
\begin{equation}
\label{eqn:var}
L_{var} = \frac{1}{C}\sum_{c=1}^C\frac{1}{N_c}\sum_{i=1}^{N_c}[||\mu_c-e_i||-\delta_v]_+^2,
\end{equation}
and push away embeddings from different instances by using:
\begin{equation}
\label{eqn:dist}
L_{dist} = \frac{1}{C(C-1)}\sum_{c_A=1}^C\sum_{c_B=1}^C[2\delta_d - ||\mu_{c_A}-\mu_{c_B}||]^2_+.
\end{equation}
Here $C$ is the number of instances, $N_c$ is the number of points in instance $c$, $\mu_c$ is the mean embedding of instance $c$, and threshold $\delta_v$ is set to be smaller than $\delta_d$. $[x]_+$ is $x$ if $x > 0$ or 0 otherwise.

At the inference stage, we use the embeddings and offsets as similarity metrics to perform mean-shift clustering to group points into instances. The similarity between point $i$ and $j$ is defined as:
\begin{equation}
    S_{ij} = \text{exp}(-(\frac{||e_i-e_j||}{\sigma_e})^2 - (\frac{||(p_i+o_i)-(p_j+o_j)||}{\sigma_p})^2),
\end{equation}
where $p_i$ denotes the position vector of point $i$, $\sigma_e$ and $\sigma_p$ are learned to automatically weight these two terms as in \cite{Han2020OccuSegO3}.

\subsection{Weakly-Supervised Approach}
\label{sec:weak}

% \subsubsection{From full to weak supervision}
% The metric learning is not directly applicable in our weak annotation setting because since only one point is annotated per instance, it is not possible to reduce the distances of embeddings from the same instance in Eqn.~\ref{eqn:var}. Thus we first expand our labels.
The metric learning is not directly applicable in our weak annotation setting. This is because since only one point is annotated per instance, it is infeasible to reduce the distances of embeddings from the same instance in Eqn.~\ref{eqn:var}. Thus, we need to expand our labels first.

We perform supervoxel partition using low-level features to segment the whole scene into small supervoxels. Points in the same supervoxel can be considered to belong to the same instance. This enables us to expand the weak point labels to supervoxel-level.

The embeddings can then be trained using Eqn.~\ref{eqn:var} and \ref{eqn:dist} on the labeled points. $\mu_c$ in the current case denotes the mean embedding of labeled points in instance $c$. The offset vectors, however, can not be learned using weak labels. Therefore, at the inference stage, only the embeddings are used to group points to instances. The similarity is defined as:
% \vspace{-5pt}
\begin{equation}
\label{eqn:similarity_weak}
    S_{ij} = \text{exp}(-(||e_i-e_j||)^2).
\end{equation}

Using only weak labels for training produces poor performance as shown in the Tab.~\ref{tab:ablation_version}. To fully utilize the unlabeled points, we propose a training framework that generates pseudo labels during training.

\myparagraph{Baseline version.} 
%Classical self-training methods first train the model on weak labels, then generate pseudo labels by model inference on the unlabeled data. Afterward, they train on both the weak labels and the pseudo labels that have high confidence. 
We first train the embeddings by metric learning on the weak labels for several epochs as described above. Then we generate pseudo instance labels by using the model to perform inference on the training set, that is, grouping points to form instances by the predicted embeddings. After that, we continue to train our model using the pseudo labels with high confidence. Specifically,  Eqn.~\ref{eqn:var} is only performed on point $i$ that satisfies $||\mu_c-e_i|| < \alpha$. Eqn.~\ref{eqn:dist} is only performed on two instances $c_A$ and $c_B$ if they satisfy $||\mu_{c_A}-\mu_{c_B}|| > \beta$. The training and pseudo label generation iterate several times. We denote this method as the baseline version.

The network trained in this way could utilize the unlabeled data, which boosts the model performance as shown in Tab.~\ref{tab:ablation_version} and Appendix. However, it still largely falls behind the fully-supervised baseline.
% In the next, we describe how we improve our naive self-training method.

\myparagraph{Click annotation version.}
The baseline version generates pseudo labels by directly using the inference way of the model, that is, performing mean-shift to group points by the network output embeddings. However, it fails to utilize the original annotation information. 
We annotate one point per instance, which means every point in the scene belongs to one of the annotated points. 
Thus the generation of pseudo instance labels in each scene can be treated as a per-point classification problem that assigns each unlabeled point to one of the annotated points, instead of directly clustering. 
This can be achieved by nearest neighbor search using the similarity metrics.

However, we take a step further by using k-means clustering with initial seeds being the annotated points. Our intuition lies in that, some annotated points may locate far from the center of their belonging instance in feature space, making the unlabeled points difficult to match them correctly. With the moving of seeds in k-means clustering, this problem can be alleviated. Each cluster contains only one annotated point, otherwise the k-means will stop. The process is visualized in Fig.~\ref{fig:overview}.
%However, some labeled supervoxels are far from their instances center, making unlabeled points hard to be correctly classified. Although our well-defined similarity metrics partially addressed this problem
%Inspired by k-means clustering, instead of simply assigning the labels once, we propose to perform this process multiple times, with every time we move the 

\begin{figure}[h]
\vspace{-10pt}
\centering
\includegraphics[width=0.35\textwidth, angle=0]{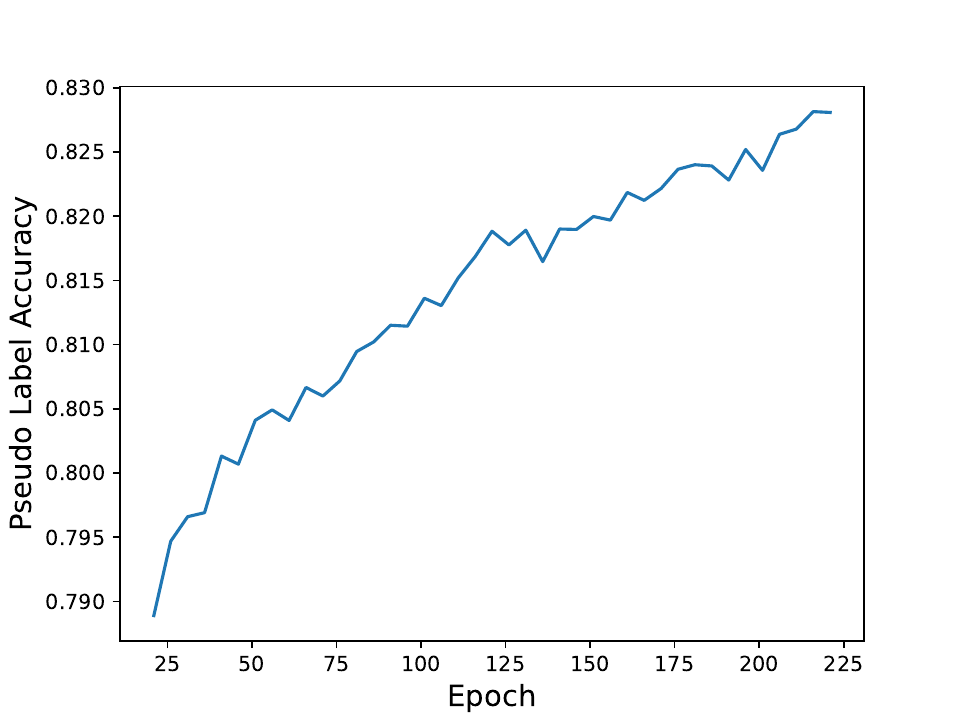}
\caption{Pseudo label generation accuracy in different training epochs.}
\label{fig:match_acc}

\end{figure}

We generate pseudo instance labels for all unlabeled points. This enables us to learn the offset vector described in the fully-supervised counterpart.
The network is trained on all pseudo instance labels without thresholding. We empirically find that the pseudo label generation accuracy increases with the training goes (see Fig.~\ref{fig:match_acc}), thus we regenerate the pseudo labels for each input scene during training. 

We also generate pseudo semantic labels. For an unlabeled point $i$, we use the semantic label of its matched annotated point $j$ as the pseudo label. The network is trained only on point $i$ that satisfies $S_{ij} >= \theta$. This is because for semantic segmentation, only a small number of labels is needed for training to achieve a high accuracy \cite{hu2021sqn}, and we find that more false pseudo labels will have negative effects on the accuracy. The whole training process is depicted in Alg.~\ref{alg:train}. We denote this method as the click annotation version.

%For semantic segmentation, only a small amount of labels are needed for training to achieve a high accuracy \cite{hu2021sqn}, and we find that the additional false pseudo semantic labels will have negative effect on the accuracy.

%For an unlabeled point, instead of directly using the semantic label of its matched instance as the pseudo label, we fuse the semantic labels of all annotated points, according to the similarities, as shown in the following formula:
%\begin{equation}
%    \hat{s_i} = \frac{\sum_{j=1}^{C}S_{ij} * \overline{s_j}}{\sum_{j=1}^{C}S_{ij}},
%\end{equation}
%where $\hat{s_i}$ is the fused semantic probability for unlabeled point $i$, $\overline{s_j}$ is a one-hot vector denoting the ground truth semantic of the labeled points $j$. The pseudo semantic label of point $i$ is assigned to be $\mathop{argmax} \hat{s_i}$ if $\mathop{max} \hat{s_i} > \theta$.

% Different from the pseudo instance label generation, we only assign pseudo semantic label for unlabeled point $i$ if $\hat{s_i} > \theta$. This is based on two reasons.
% Firstly, the semantic segmentation only requires a small amount of labels for training to achieve a high accuracy \cite{hu2021sqn}. Secondly, we find that the additional false pseudo semantic labels will have negative effects on the accuracy.
\vspace{-3pt}
\begin{algorithm}[h]

\caption{Training process of the click annotation version. $\text{PSEUDO\_GEN}$ denotes the pseudo label generation process. $s, e, o$ are semantic probabilities, embeddings, and offsets, respectively. $t$ is a hyper-parameter.} \label{algorithm}
\KwIn{A training sample $(x, y_1, y_2)$, denoting the input scene, the click-level instance and semantic labels.}
\KwOut{Training loss $L$.}
 $s, e, o \leftarrow \text{NETWORK}(x)$\\
 $\overline{y_1} \leftarrow$ expand $y_1$ by supervoxel\\
 $\overline{y_2} \leftarrow$ expand $y_2$ by supervoxel\\
 $L \leftarrow L_{var}(e, \overline{y_1}) + L_{dist}(e, \overline{y_1}) + L_{CE}( s, \overline{y_2})$\\
\If{\text{current training epoch} $\geq t$}{
$\hat{y_1}, \hat{y_2} \leftarrow \text{PSEUDO\_GEN}
(s, e, o, x, y_1) $\\
Filter out unreliable labels in $\hat{y_2}$ \\
$L\leftarrow L + L_{var}(e, \hat{y_1}) + L_{dist}(e, \hat{y_1}) + L_{CE}( s, \hat{y_2}) + L_{regress}(o, \hat{y_1})$\\
}
\label{alg:train}
\vspace{-0pt}
\end{algorithm}

\myparagraph{Similarity.} We modify the similarity metrics to better adapt to weak supervision setting, thus helping to generate more accurate pseudo labels. The new similarity formula between point $i$ and $j$ is defined as:
\begin{equation}
\label{eqn:similarity}
    S_{ij} = Q_{ij} * \text{exp}(-(\frac{D^e_{ij}}{\sigma_e})^2 - (\frac{D^p_{ij}}{\sigma_p})^2),
\end{equation}
which contains three terms: the embedding distance $D^e_{ij}$, the spatial distance $D^p_{ij}$, and the semantic similarity $Q_{ij}$. The embedding distance $D^e_{ij}$ is simply $||e_i-e_j||$ which is the same as Eqn.~\ref{eqn:similarity_weak}.

The semantic similarity score $Q_{ij}$ between two points is calculated by: $Q_{ij} = \frac{s_i\cdot s_j}{||s_i||||s_j||}.$
% \begin{equation}
%     Q_{ij} = \frac{s_i\cdot s_j}{||s_i||||s_j||}.
% \end{equation}
We take this term into account for the following reason:
the embeddings trained on weak supervoxel-level labels can only capture local similarities by low-level features like colors, normals, etc., depending on the supervoxel partition method. If an instance is composed of several parts, with each part having large differences in their ``looks'', simply using the embeddings may not be able to match them correctly. 
% Unlike the embeddings, even with only click-level weak labels used for training, 
In contrast,
the prediction of semantics is relatively robust for different parts of an object. 
For example, the chair leg can be correctly classified as \textit{chair} in Fig.~\ref{fig:match_error}, while it can not be recognized as part of the chair in the instance embedding space. 
This is because although the number of semantic labels of \textit{chair} is limited in one scene, there are still many labels of \textit{chair} in the whole dataset. The learning on these labels helps the network to recognize the different parts of the chair.
% Thus, since the semantic can be well predicted, and p
Points belonging to the same instance should share the same semantic labels, so we add this term to help better match them.

\begin{figure}[t]

\centering
\includegraphics[width=0.38\textwidth, angle=0]{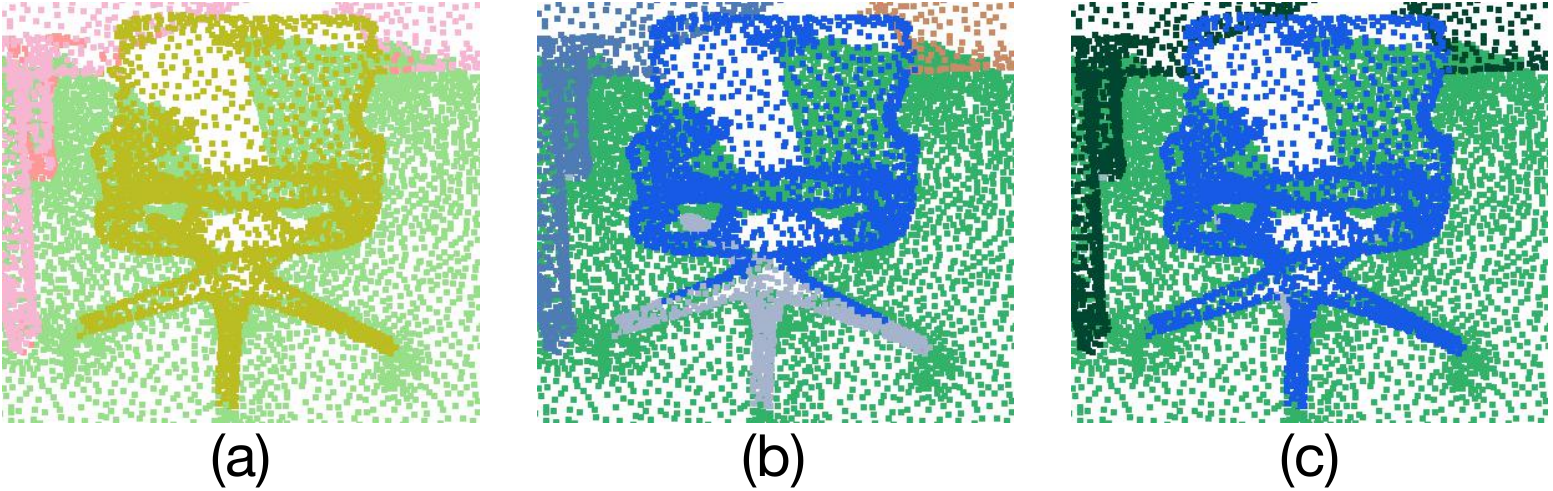}
\caption{A chair in the training set. (a) the predicted semantics, (b)(c) the pseudo instance label generated without and with using semantics.}
\label{fig:match_error}
\vspace{-10pt}
\end{figure}

The spatial distance $D^p_{ij}$ is defined as $||p_i-p_j||$, where $p_i$ denotes the position vector of point $i$. This is based on the observation that points belonging to the same instance are often spatially adjacent. Adding this term avoids some false cases that assign unlabeled points to the annotated points that have the same semantic labels and similar local structures but belong to different instances. After some training epochs on the pseudo labels, the offset vectors are learned, we can then calculate $D^p_{ij}$ by $||(p_i+o_i)-(p_j+o_j)||$, same as the fully supervised baseline method.

\myparagraph{One-thing-multi-click.} The current method can also be extended to the annotation setting of one-thing-multi-click, which means annotating multiple points per instance. The pseudo label generation way is a bit different than that of one-thing-one-click.
%we do not perform k-means clustering since some annotated points may belong to the same instance.
%The later directly assigns each unlabeled points to one of the annotated points, since every annotated point belongs to a different instance.
In the multi-click case, the k-means is not applicable. Thus we directly assign unlabeled points to instances by nearest neighbor search. We calculate the likelihood of an unlabeled point belonging to an instance by the mean of similarities between this point and all the annotated points of that instance. Then we simply choose the most likely instance the point belongs to as its pseudo label.

\subsection{Inference}
At inference stage, for instance segmentation, we perform mean-shift clustering using the similarity metrics to get the predicted instances. For semantic segmentation, we perform a fusion step for each point to fuse the semantic probabilities of other points according to the similarities. The insight is that for similar points, they should share the same semantic labels. The calculation formula is:
\begin{equation}
    \hat{s_i} = \frac{\sum_{j=1}^{N}\mathbb{I}_{ij}*S_{ij} * s_j}{\sum_{j=1}^{N}\mathbb{I}_{ij}*S_{ij}},
\end{equation}
where $\hat{s_i}$ denotes the fused semantic probability of point $i$, $N$ is the number of points in the scene, $\mathbb{I}_{ij}$ is $1$ if $S_{ij}>\gamma$ and $0$ otherwise. The final prediction for point $i$ is $\mathop{argmax} \hat{s_i}$. This helps to smooth the predictions of similar points, resulting in a more consistent semantic map.
\label{sec:inference}

% In the extremely limited annotation setting that only one point is labeled per instance, we could not directly apply current point cloud instance segmentation method. Thus we first generate pseudo labels for the unlabeled points. Given such annotation setting, every points must belong to one of the labeled points (there are very few points that do not belong to any instance, which can be ignored). Thus, we treat the pseudo generation problem as a classification problem that assign each unlabeled points to one of the labeled points. We learn the similarities between points, and use k-means clustering to classify the unlabels. We first describe the similarity 

% \begin{figure}[t]
% \centering
% \includegraphics[width=\textwidth, angle=0]{figs/pseudo.pdf}
% \caption{Pseudo label visualization. The generated pseudo instance labels in epoch \#200 have better quality than that in epoch \#10.}
% \label{fig:qualitative_pseudo}
% \vspace{-10pt}
% \end{figure}

\begin{figure*}[t]
\centering
\includegraphics[width=\textwidth, angle=0]{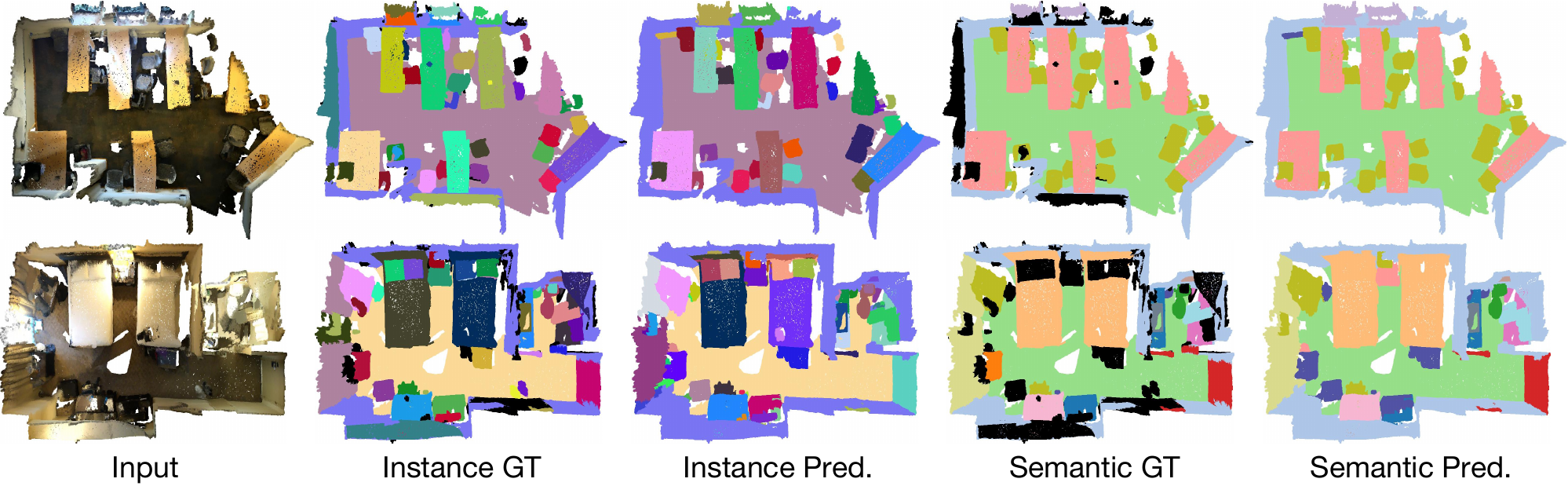}
\caption{Visualization of the semantic and instance results of our click annotation version on the ScanNetV2 validation set. }
\label{fig:qualitative}
% \vspace{-10pt}
\end{figure*}

% \begin{figure*}[h]
% \centering
% % \vspace{-5pt}
% \includegraphics[width=\textwidth, angle=0]{figs/pseudo.pdf}
% \caption{Pseudo label visualization. The generated pseudo instance labels in epoch \#200 have better quality than that in epoch \#10, because the training and pseudo label generation benefit from each other. This differs from Seggroup\cite{tao2022seggroup}, which separates the two processes.}
% \label{fig:qualitative_pseudo}

% \end{figure*}

\begin{figure*}[t]
\centering
\includegraphics[width=\textwidth, angle=0]{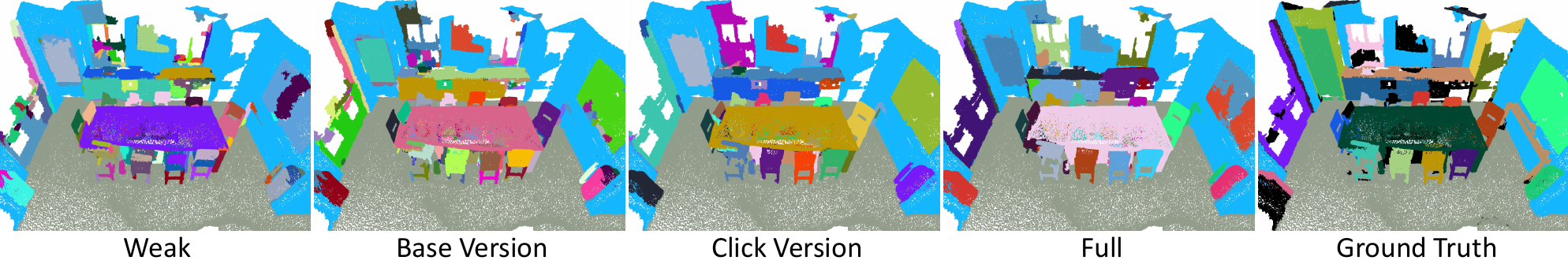}

\caption{Qualitative results of different versions of our method on the ScanNetV2 validation set. ``Weak'' denotes training only on weak labels. ``Base Version'' denotes the baseline version of our method. ``Click Version'' denotes the click annotation version. ``Full'' denotes the fully-supervised counterpart method.}
\label{fig:qualitative_version}
\end{figure*}

\begin{table*}[t]
\vspace{15pt}
\centering

\setlength{\tabcolsep}{2.5mm}{
\begin{tabular}{c c c c}
\Xhline{3\arrayrulewidth}
\multicolumn{4}{c}{\textbf{(a) Instance Segmentation on ScanNetv2}} \\
Methods  & Supervision & mAP@50 (\%) (val split)    &  mAP@50 (\%) (test split) \\ \hline
SGPN \cite{wang2018sgpn} & 100\% & - & 14.3 \\ 
3D-SIS \cite{Hou20193DSIS3S} & 100\% & 22.5 & 38.2 \\ 
3D-BoNet \cite{yang2019learning} & 100\% & - & 48.8 \\ 
MASC \cite{liu2019masc} & 100\% & - & 44.7 \\
MTML \cite{Lahoud20193DIS} & 100\% & 40.2 & 54.9 \\ 
PointGroup \cite{Jiang2020PointGroupDP} & 100\% & 56.9 & 63.6 \\
Occuseg \cite{Han2020OccuSegO3} & 100\% & 60.7 & 67.2 \\ 
% SSTNet \cite{liang2021instance} & 100\% & 64.7 & 69.8 \\ 
% HAIS \cite{chen2021hierarchical} & 100\% & 64.1 & 69.9 \\ 
\hline
Fully-sup baseline & 100\% & 57.7 & - \\ Weakly-sup baseline & 1t1c (0.02\%) & 25.5 & -\\

\hline
SegGroup \cite{tao2022seggroup} & Supervoxels & 43.4 & 44.5\\ \hline
ClickSeg & 1t1c (0.02\%) & 51.9 & 53.9 \\
ClickSeg & 1t2c (0.04\%) & 55.2 & - \\
ClickSeg & 1t3c (0.06\%) & 57.0 & - \\
\Xhline{3\arrayrulewidth}
\multicolumn{4}{c}{\textbf{(b) Semantic Segmentation on ScanNetv2}} \\
Methods  & Supervision & mIoU (\%) (val split)    &  mIoU (\%) (test split) \\ \hline
PointNet \cite{Qi2017PointNetDH} & 100\% &53.5 & 55.7 \\ 
PointConv \cite{Wu2019PointConvDC} & 100\% &61.0 & 66.6 \\ 
KPConv \cite{Thomas2019KPConvFA} & 100\% & 69.2 & 68.4 \\ 
SparseConvNet \cite{Graham20183DSS} & 100\% &69.3 & 72.5 \\ 
MinkowskiNet \cite{choy20194d} & 100\% &72.2 & 73.6 \\ 
Occuseg \cite{Han2020OccuSegO3} & 100\% & - & 76.4 \\ \hline
Fully-sup baseline & 100\% & 72.0 & - \\ 
Fully-sup baseline* & 100\% & 74.6 & - \\ 
Weakly-sup baseline & 1t1c (0.02\%) & 61.9 & -\\
\hline
MPRM \cite{wei2020multi} & Subcloud & 46.6 & 41.1 \\
SegGroup \cite{tao2022seggroup} & Supervoxels & 62.4 & 61.1\\
% SQN \cite{hu2021sqn} & 0.01\% & - & 35.9 \\ 
SQN \cite{hu2021sqn} & 0.1\% & - & 51.6 \\ 
OTOC \cite{liu2021one} & 1t1c (0.02\%) & 70.5 & 69.1 \\ \hline
ClickSeg & 1t1c (0.02\%) & 71.4 & 70.3 \\
ClickSeg & 1t2c (0.04\%) & 72.6 & - \\
ClickSeg & 1t3c (0.06\%) & 74.1 & - \\

\Xhline{3\arrayrulewidth}

\end{tabular}
}
\vspace{10pt}
\caption{Instance and semantic segmentation results on ScanNetV2 validation and test set. ``1t1c'' means annotating one arbitrary point per instance. Seggroup \cite{tao2022seggroup} annotates the most central supervoxel of each instance. ``Fully-sup baseline*'' means the fully-supervised baseline method with the semantic fusion strategy described in Sec.~\ref{sec:inference}. ``Weakly-sup baseline'' means training only on weak supervoxel-level labels as described in Sec.~\ref{sec:weak}.
Due to the ScanNetV2 submission policy, we only report the test set result of ClickSeg under the ``1t1c'' setting. }
\label{tab:scannet_instance}
\end{table*}

\section{Experiments}
\label{sec:experiment}

\subsection{Experimental Setup}
\myparagraph{Datasets and metrics.} We evaluate ClickSeg on two popular datasets, ScanNetV2 \cite{Dai2017ScanNetR3} and S3DIS \cite{armeni2017joint}. The ScanNetV2 dataset contains 1513 scenes that have publicly available labels and 100 scenes for hidden testing. These 1513 scenes are further split into 1201 and 312 scenes officially for training and validation. We follow this split and evaluate on both validation and testing sets. The S3DIS dataset includes 6 large-scale areas. Each area contains dozens of indoor scenes. We follow other methods \cite{liu2021one, Jiang2020PointGroupDP} to choose Area-5 as the testing set, and other areas as the training set. We evaluate both semantic and instance results. The evaluation metric for semantic segmentation is mean Intersection-over-Union (mIoU), and for instance segmentation is mean Average Precision (mAP).

\myparagraph{Annotation details.} We annotate one single arbitrary point for every object that appears in the scene to indicate its semantic class, which is the same as \cite{liu2021one}. The instance id is naturally contained. There are on average 31.93 points per scene in ScanNetV2 training set and 36.7 points per scene in S3DIS training set are annotated under this setting. The average number of points per scene in ScanNetV2 training set is 145,170.8, and 955,975.5 in S3DIS. The annotation ratio is only about 0.02\% in ScanNetV2 and 0.004\% in S3DIS.

\myparagraph{Implementation details.}
We use the Adam optimizer with a learning rate of 5e-3. The batch size is 8 for ScanNetV2 and 4 for S3DIS. A single Tesla V100 (32GB) GPU is used for training. We generate the supervoxels by using the graph-based method \cite{felzenszwalb2004efficient} for ScanNetV2, and SSP \cite{landrieu2018large} for S3DIS. Following other methods \cite{Jiang2020PointGroupDP, liang2021instance}, we pre-process scenes in S3DIS by randomly sub-sample points at a rate of 1/4 since the point density is too high. The voxel size is set to 0.02m. We first train the network using expanded supervoxel-level weak labels for 250 epochs. Then we generate pseudo labels and train using both the pseudo labels and the weak labels for another 250 epochs. We set the hyper-parameters $\delta_v = 0.2$, $\delta_d = 1.5$, $\alpha=0.6$, $\beta=1.5$, $\theta=0.9$, $\gamma=0.3$.

% \begin{table}[t]
% \centering
% \caption{Semantic results on ScanNetV2 validation and test set. ``Fully-sup baseline*'' means the fully-supervised baseline method with semantic fusion strategy described in Sec.~\ref{sec:inference}.}
% \label{tab:scannet_semantic}
% \small
% \setlength{\tabcolsep}{2.5mm}{
% \begin{tabular}{c c c c}
% \Xhline{3\arrayrulewidth}
% Methods  & Supervision & mIoU (\%) (val split)    &  mIoU (\%) (test split) \\ \hline
% % PointNet \cite{Qi2017PointNetDH} & 100\% &53.5 & 55.7 \\ 
% % PointConv \cite{Wu2019PointConvDC} & 100\% &61.0 & 66.6 \\ 
% KPConv \cite{Thomas2019KPConvFA} & 100\% & 69.2 & 68.4 \\ 
% SparseConvNet \cite{Graham20183DSS} & 100\% &69.3 & 72.5 \\ 
% MinkowskiNet \cite{choy20194d} & 100\% &72.2 & 73.6 \\ 
% Occuseg \cite{Han2020OccuSegO3} & 100\% & - & 76.4 \\ \hline
% Fully-sup baseline & 100\% & 72.0 & - \\ 
% Fully-sup baseline* & 100\% & 74.6 & - \\ \hline
% MPRM \cite{wei2020multi} & Subcloud & 46.6 & 41.1 \\
% SegGroup \cite{tao2020seggroup} & Supervoxels & 62.4 & 61.1\\
% % SQN \cite{hu2021sqn} & 0.01\% & - & 35.9 \\ 
% SQN \cite{hu2021sqn} & 0.1\% & - & 51.6 \\ 
% OTOC \cite{liu2021one} & 1t1c (0.02\%) & 70.5 & 69.1 \\ \hline
% Ours & 1t1c (0.02\%) & 71.4 & 70.3 \\
% Ours & 1t2c (0.04\%) & 72.6 & - \\
% Ours & 1t3c (0.06\%) & 74.1 & - \\
% \Xhline{3\arrayrulewidth}
% \end{tabular}
% }
% % \vspace{-10pt}
% \end{table}

\subsection{Results}

\begin{table}[t]
\centering

\begin{tabular}{c c c}
\Xhline{3\arrayrulewidth}
\multicolumn{3}{c}{\textbf{(a) Instance Segmentation on S3DIS}} \\
Methods & Supervision & mAP@50 (\%) \\ \hline
PointGroup \cite{Jiang2020PointGroupDP} & 100\% & 57.8 \\
SSTNet \cite{liang2021instance} & 100\% & 59.3 \\ \hline
Fully-sup baseline & 100\% & 53.4\\ 
Weakly-sup baseline & 1t1c (0.004\%) & 21.0\\ \hline
ClickSeg & 1t1c (0.004\%) & 45.7 \\
\Xhline{3\arrayrulewidth}
\multicolumn{3}{c}{\textbf{(b) Semantic Segmentation on S3DIS}} \\
Methods  & Supervision & mIoU (\%) \\ \hline
PointNet \cite{Qi2017PointNetDH} & 100\% & 41.1\\
TangentConv \cite{tatarchenko2018tangent} & 100\% & 52.8 \\
MinkowskiNet\cite{Choy20194DSC} & 100\% & 65.4\\ 
KPConv \cite{Thomas2019KPConvFA} & 100\% & 67.1\\ 
\hline
Fully-sup baseline & 100\% & 65.5 \\ 
Weakly-sup baseline & 1t1c (0.004\%) & 51.0\\ \hline
SQN \cite{hu2021sqn} & 0.01\% & 45.3\\
OTOC \cite{liu2021one} & 1t1c (0.004\%) & 50.2\\\hline
ClickSeg & 1t1c (0.004\%) & 57.2 \\
\Xhline{3\arrayrulewidth}

\end{tabular}
\vspace{5pt}
	\caption{Instance and semantic results on S3DIS Area-5.}
 \label{tab:s3dis_instance}

\end{table}

% \begin{table}[t]
% \caption{Semantic results on S3DIS. We report mIoU(\%) on Area-5.}
% \label{tab:s3dis_semantic}
% \centering
% \small
% \begin{tabular}{c c c}
% \Xhline{3\arrayrulewidth}
% Methods  & Supervision & mIoU \\ \hline
% PointNet \cite{Qi2017PointNetDH} & 100\% & 41.1\\
% % TangentConv \cite{tatarchenko2018tangent} & 100\% & 52.8 \\
% MinkowskiNet \cite{choy20194d} & 100\% & 65.4\\ 
% %KPConv \cite{Thomas2019KPConvFA} & 100\% & 67.1\\ 
% \hline
% Fully-sup baseline & 100\% & 65.5 \\ \hline
% SQN \cite{hu2021sqn} & 0.01\% & 45.3\\
% OTOC \cite{liu2021one} & 1t1c (0.004\%) & 50.2\\\hline
% ClickSeg & 1t1c (0.004\%) & 57.2 \\
% \Xhline{3\arrayrulewidth}
% \end{tabular}
% \end{table}

\myparagraph{ScanNetV2.}
We compare ClickSeg with fully-supervised methods and other weakly or semi-supervised methods.
The instance and the semantic segmentation results are shown in Tab.~\ref{tab:scannet_instance}.
It achieves about 95\% mIoU and 90\% mAP@50 of the fully-supervised baseline using only 0.02\% labels for training. Meanwhile, ClickSeg also surpass all other weakly supervised methods in both semantic and instance results. Notice that SegGroup \cite{tao2022seggroup} requires the annotator to annotate the most central supervoxel for each instance, while our method only requires to annotate one arbitrary point for each instance. 
% while our method only requires to annotate one single random point for each instance. 
The qualitative results are shown in Fig~.\ref{fig:qualitative}. %and Fig~.\ref{fig:qualitative_pseudo}.

We additionally report the results of ClickSeg under the annotation setting of one-thing-two/three-click in Tab.~\ref{tab:scannet_instance}. With each instance annotating up to three points, it even achieves about 99\% accuracy of our fully-supervised baseline in terms of both semantic and instance segmentation.

\myparagraph{S3DIS.} We report the semantic and instance segmentation results of ClickSeg on S3DIS Area-5. Since no other weakly or semi-supervised 3D instance segmentation method have tested on S3DIS, we only compare with fully-supervised methods and other weakly or semi-supervised semantic segmentation methods. The results are shown in Tab.~\ref{tab:s3dis_instance}, the semantic result of ClickSeg outperforms other weakly supervised methods. It surpasses \cite{liu2021one} by 7\% of mIOU using the same annotations. It also achieves 85.5\% instance mAP@50 compared with the fully-supervised baseline by using only 0.004\% labels for training.

% \begin{table}[t]
% \centering
% \caption{Ablation study on different versions. ``Weak'' denotes training only on weak labels. ``Base'' denotes the base self-training method. ``Improved'' denotes the improved version. ``Full'' denotes the fully-supervised baseline method.}
% \label{tab:ablation_version}
% \setlength{\tabcolsep}{5mm}{
% \begin{tabular}{c | c c c c}
% \Xhline{3\arrayrulewidth}
%  & Weak  & Base & Improved & Full\\ \hline
%  mAP@50 & 25.5 & 34.5 & 51.9 & 57.7\\
 
% \Xhline{3\arrayrulewidth}
% \end{tabular}
% }
% \end{table}

\subsection{Ablation Study}
\label{sec:ablation}
We test the effectiveness of different components in ClickSeg. The ablation studies are conducted on ScanNetV2 validation set, under the one-thing-one-click annotation setting.

\begin{table}[t]
	\centering
		\centering
		\small
\setlength{\tabcolsep}{1.5mm}{
\begin{tabular}{c | c c c c}
\Xhline{3\arrayrulewidth}
 & Weak  & Base & Click & Full\\ \hline
 mAP@50 & 25.5 & 34.5 & 51.9 & 57.7\\
\Xhline{3\arrayrulewidth}
\end{tabular}
}
\vspace{5pt}
		\caption{Instance segmentation results of different versions of our method. ``Weak'' denotes training only on weak labels. ``Base'' denotes the baseline version. ``Click'' denotes the click annotation version. ``Full'' denotes the fully-supervised counterpart method.}
\label{tab:ablation_version}
\end{table}

\subsubsection{Ablation on Instance Segmentation}

\label{sec:ablation_instance}

\myparagraph{Different versions.} We show the instance mAP@50 results of different versions of ClickSeg in Tab.~\ref{tab:ablation_version} and the qualitative results in Fig.~\ref{fig:qualitative_version}. The network trained directly on weak supervoxel-level labels only achieves 25.5\% mAP@50. This is because the embeddings can only capture local similarity, thus resulting in over-segmentation.
The baseline version utilizes the unlabeled data for training, which improves the results by 9.0\%. However, the pseudo labels are not accurate enough, and the high threshold filters out most of the pseudo labels. Thus the accuracy still largely falls behind the fully-supervised method. The click annotation version use a different way of pseudo label generation, and proposes some modifications to the similarity metrics, which greatly increase the pseudo label accuracy. The final results outperform the baseline version by a significant increase of 17.4\% mAP@50, which validates the effectiveness of our method.

\myparagraph{Effect of k-means.} In the click annotation version, instead of using nearest neighbor search to assign pseudo labels, the k-means helps to alleviate the problem of misclassification. Because some annotated points may locate at the border of their belonging instances in the feature space, unlabeled points can be misclassified. The k-means gradually moves the seeds to help to correct these false cases. Tab.~\ref{tab:ablation_k_means} shows using k-means increases the accuracy by 1.4\% mAP@50.

\myparagraph{Effect of different similarity metrics.}
We test the effect of different designs of similarity metrics. There are three terms in the similarity calculation formula: the embedding distance, the spatial distance, and the semantic similarity. These terms help generate more accurate pseudo labels. As shown in Tab.~\ref{tab:ablation_metrics}, we remove some of these terms to test their effects. The results show that they all improve the instance accuracy in terms of mAP@50, thus proving the effectiveness of our designs.

% \begin{table}[t]
% \centering
% \caption{Ablation study on semantic segmentation.}
% \label{tab:ablation_semantic}
% \setlength{\tabcolsep}{3mm}{
% \begin{tabular}{c c c |c}
% \Xhline{3\arrayrulewidth}
% Pseudo instance labels  & Pseudo semantic labels & Semantic fusion & mIoU \\ \hline
%  & & & 61.9 \\
%  \checkmark & & & 66.1 \\
%  \checkmark & \checkmark & & 67.9 \\
% %  \checkmark &  & \checkmark & 70.6 \\
%  \checkmark & \checkmark & \checkmark & 71.4 \\
% \Xhline{3\arrayrulewidth}
% \end{tabular}
% }
% \end{table}

\begin{table}[t]
\centering
\small
\setlength{\tabcolsep}{1.5mm}{
\begin{tabular}{c | c c}
\Xhline{3\arrayrulewidth}
 & Nearest neighbor  & K-means\\ \hline
 mAP@50 & 50.5 & 51.9 \\
\Xhline{3\arrayrulewidth}
\end{tabular}
}
\vspace{5pt}
\caption{Results of different way of generating pseudo labels in the click annotation version.}
\label{tab:ablation_k_means}

\end{table}

\begin{table}[t]
	\centering

\setlength{\tabcolsep}{1.5mm}{
\begin{tabular}{c c c | c}
\Xhline{3\arrayrulewidth}
Embedding  & Spatial & Semantic & mAP@50   \\ \hline
\checkmark &   &  &  48.7        \\
\checkmark & \checkmark  & & 49.2   \\ 
\checkmark &  & \checkmark &   50.7   \\ 
\checkmark & \checkmark & \checkmark &  51.9   \\

\Xhline{3\arrayrulewidth}
\end{tabular}
}
\vspace{5pt}
		\caption{Effects of different terms in the similarity formula.}
\label{tab:ablation_metrics}
\end{table}

\begin{table}[t]
		
		\centering
		
\setlength{\tabcolsep}{1.5mm}{
\begin{tabular}{c c c |c}
\Xhline{3\arrayrulewidth}
Pseudo Inst.  & Pseudo Sem. & Fusion & mIoU \\ \hline
 & & & 61.9 \\
 \checkmark & & & 66.1 \\
 \checkmark & \checkmark & & 67.9 \\
%  \checkmark &  & \checkmark & 70.6 \\
 \checkmark & \checkmark & \checkmark & 71.4 \\
\Xhline{3\arrayrulewidth}
\end{tabular}
}

\vspace{5pt}

		\caption{Ablation study on semantic segmentation.}
  \label{tab:ablation_semantic}
% 	\vspace{-10pt}
\end{table}

\subsubsection{Ablation on Semantic Segmentation}

The ablation results on semantic segmentation are shown in Tab.~\ref{tab:ablation_semantic}. The model trained on supervoxel-level weak labels achieved 61.9\% mIoU. We find that even only with the training on instance pseudo labels, the semantic segmentation accuracy increased by 4.2\% as well, which is shown in the second raw of Tab.~\ref{tab:ablation_semantic}. This is because we use a multi-task framework, where each task can benefit from the others. The training on the semantic pseudo labels further boosted the results by 1.8\% mIoU. At inference stage, for each point, we fuse the semantic probabilities of similar other points. This utilizes the instance information to guide the semantic segmentation, which helps to smooth the predictions of similar points, with an increase of 3.5\% mIoU.

\section{Conclusion and Discussion}
\label{sec:discussion}
In this paper, we present ClickSeg. We first develop a baseline training method for weakly supervised instance segmentation. To utilize the property of the click-level annotation setting, we further propose a specific training framework termed click annotation version. Extensive experiments prove the superior performance of our method. ClickSeg also has limitations. The baseline version could be applied to some other annotation settings but is not accurate enough compare with fully supervised methods. The click annotation version has higher accuracy but may have performance degradation problem when not all instances are annotated for at least one point. Setting threshold when generating pseudo labels may help to tackle this problem. %There are also failure cases, e.g, objects that composed of several parts of different shapes tend to be over-segmented under the one-thing-one-click annotation setting. Although we have alleviated it by modifying similarity metrics, it still happens because the embeddings may only capture local similarities. With more than one point annotated per instance, this problem can be largely addressed. 

{\small
\bibliographystyle{ieee_fullname}
\bibliography{egbib}
}

\end{document}